\documentclass[letterpaper]{article} 
\usepackage{aaai24}  
\usepackage{times}  
\usepackage{helvet}  
\usepackage{courier}  
\usepackage[hyphens]{url}  
\usepackage{graphicx} 
\usepackage{natbib}  
\usepackage{caption} 
\usepackage{algorithm}
\usepackage{algorithmic}

\usepackage{newfloat}
\usepackage{listings}
\DeclareCaptionStyle{ruled}{labelfont=normalfont,labelsep=colon,strut=off} 
\floatstyle{ruled}
\newfloat{listing}{tb}{lst}{}
\floatname{listing}{Listing}
\usepackage[utf8]{inputenc} 
\usepackage[T1]{fontenc}    
\usepackage{url}            
\usepackage{booktabs}       
\usepackage{amsfonts}       
\usepackage{nicefrac}       
\usepackage{microtype}      
\usepackage{epsfig}
\usepackage{graphicx}
\usepackage{amsmath}
\usepackage{amssymb}
\usepackage{color}
\usepackage{bm}
\usepackage{multicol,multirow}
\usepackage{algorithm,algorithmic}
\usepackage{bbding}
\usepackage{arydshln}
\usepackage{latexsym}
\usepackage{tikz}
\usepackage{float}
\usepackage{comment}
\usepackage{amsmath,amssymb} 
\usepackage{color}
\usepackage{multirow}
\usepackage{pifont}

\usepackage{multirow,multicol, makecell, booktabs}
\usepackage{tabularray}
\usepackage{booktabs,dcolumn,siunitx}

\title{DF-3DFace: One-to-Many Speech Synchronized \\ 3D Face Animation with Diffusion }
\author{
    Se Jin Park, Joanna Hong, Minsu Kim, Yong Man Ro\thanks{Corresponding author.}
}
\affiliations{
Image and Video Systems Lab, KAIST\\
\{jinny960812, joanna2587, ms.k, ymro\}@kaist.ac.kr

}

\begin{document}

\maketitle

\begin{abstract}
Speech-driven 3D facial animation has gained significant attention for its ability to create realistic and expressive facial animations in 3D space based on speech. Learning-based methods have shown promising progress in achieving accurate facial motion synchronized with speech. However, one-to-many nature of speech-to-3D facial synthesis has not been fully explored: while the lip accurately synchronizes with the speech content, other facial attributes beyond speech-related motions are variable with respect to the speech. To account for the potential variance in the facial attributes within a single speech, we propose DF-3DFace, a diffusion-driven speech-to-3D face mesh synthesis. DF-3DFace captures the complex one-to-many relationships between speech and 3D face based on diffusion. It concurrently achieves aligned lip motion by exploiting audio-mesh synchronization and masked conditioning. Furthermore, the proposed method jointly models identity and pose in addition to facial motions so that it can generate 3D face animation without requiring a reference identity mesh and produce natural head poses. We contribute a new large-scale 3D facial mesh dataset, 3D-HDTF to enable the synthesis of variations in identities, poses, and facial motions of 3D face mesh. Extensive experiments demonstrate that our method successfully generates highly variable facial shapes and motions from speech and simultaneously achieves more realistic facial animation than the state-of-the-art methods. 
\end{abstract}

\section{1. Introduction}
Speech-driven 3D facial animation aims to generate facial movements that accurately reflect the corresponding speech sounds with natural and realistic facial motion. It has been gaining increasing attention for its ability to represent vivid facial movements in 3D space and its applicability to the animation industries, \textit{e.g.}, 3D games, and virtual reality, where the avatars animate in a 3D environment. 

Among the works in speech-driven 3D facial animation, learning-based methods \cite{fan2022faceformer, cudeiro2019capture, xing2023codetalker, richard2021meshtalk, cao2005expressive, fan2022joint, habibie2021learning, pham2018end, wang20213d, taylor2017deep, karras2017audio} have recently shown significant progress in achieving realistic facial motion in sync with speech. They adopt a data-driven approach, utilizing paired audio-mesh data to generate facial animation for a given speaker in a 3D vertex space. These methods employ techniques such as latent space learning \cite{richard2021meshtalk, xing2023codetalker}, audio feature extraction \cite{cudeiro2019capture}, and transformer model \cite{fan2022faceformer} and direct toward improving the accuracy of facial motion for resolving the highly ill-posed audio-visual mapping task.

Despite numerous efforts to generate accurate 3D facial motion from speech, the one-to-many nature of speech-to-facial synthesis has not been well addressed. Input speech can decide mouth movements and rough facial shape, but it cannot precisely determine facial attributes that have weak correlations with the speech \cite{xing2023codetalker, tang2022memories, ma2023styletalk}. To be specific, the same person can speak the same utterance with variations in speaking style, movement of the eyebrows, eyelids, and head pose. Yet, current works deterministically synthesize 3D face animation from speech through one-to-one mapping. Furthermore, the synthesis is mostly restricted to the inner facial motion \cite{fan2022faceformer, xing2023codetalker, karras2017audio, fan2022joint, wang20213d, richard2021meshtalk}. To elaborate, the identity component of the 3D face from the speech is not synthesized but is given at the input as an identity template mesh which is the mesh of the reference identity in neutral expression. Also, the head pose is left for manual control. 3D face datasets are mostly in neutral head pose. Existing 3D face datasets with synchronized audio \cite{eth_biwi_00760, karras2017audio, cudeiro2019capture, richard2021audio, wuu2022multiface} are limited because they are shot in a controlled environment with static poses or static expressions and contain only dozens of characters with a few minutes of utterance. Such limited scope of 3D face datasets presents a constraint in capturing diverse and natural 3D facial animations.

To tackle the aforementioned challenge, we propose DF-3DFace, a novel framework for speech-to-3D face synthesis. DF-3DFace employs a diffusion mechanism to capture the complex one-to-many relationships between speech and 3D face mesh, introducing stochasticity to the animation. It generates diverse facial animations while maintaining high-fidelity lip motion by exploiting mesh-audio synchronization and masked conditioning. We model identity and head pose in addition to facial motion to achieve more comprehensive facial animations. We show that the three facial components can not only be generated directly from speech but also be individually controlled by providing the reference at the input. 
Since existing 3D face datasets \cite{eth_biwi_00760, karras2017audio, cudeiro2019capture, richard2021audio, wuu2022multiface} lack diversity in the facial motions and shapes, we construct a new large-scale 3D facial mesh dataset, namely 3D-HDTF, from a large-scale high-resolution 2D talking face dataset \cite{zhang2021flow}. Leveraging the 3D-HDTF, we build a scalable and unified 3D face animation model that captures variations in identities, poses, and facial motions of 3D face mesh from speech. 

Our contributions are four-fold: (1) We propose one-to-many speech synchronized DF-3DFace, a novel speech-to-3D face mesh generation framework based on diffusion. Our diffusion approach enables diverse 3D facial animation from a single speech while fulfilling lip synchronization. 
(2) We model identity, head pose, and facial motion to achieve comprehensive 3D facial animation driven by speech. Each component can also be independently adjusted, providing controllability to the diverse generation.  
(3) We construct a large-scale 3D facial mesh dataset namely 3D-HDTF from a 2D talking face video dataset (HDTF) \cite{zhang2021flow} using a 3D face reconstruction technique \cite{DECA:Siggraph2021}. The 3D-HDTF facilitates the learning of diverse and natural variations of identities, poses, and facial motions.   
(4) Through extensive experiments, we demonstrate that DF-3DFace successfully captures the one-to-many relationships between speech and 3D face mesh, while achieving superior lip synchronization and realism compared to state-of-the-art methods. 

\section{2. Related Works}
\subsection{2.1. Speech-driven 3D Facial Animation}
Speech-driven 3D facial animation uses speech signals to generate realistic and expressive 3D facial movements in 3D vertex space. Previous works on speech-driven 3D facial animation have focused on learning the accurate mapping between audio and face mesh in terms of facial motion \cite{cao2005expressive, taylor2017deep, zhou2018visemenet, karras2017audio, cudeiro2019capture, richard2021meshtalk, fan2022faceformer, xing2023codetalker, zhou2018visemenet}. For example, \cite{cao2005expressive} employs Anime Graph structure and a search-based mechanism. \cite{taylor2017deep} takes a sliding window approach and a re-targeting technique to animate unseen identities. VOCA \cite{cudeiro2019capture} utilizes subject conditioning to capture individual speaking styles. Meshtalk \cite{richard2021meshtalk} disentangles audio-correlated and -uncorrelated information in a categorical latent space, and autoregressively samples an audio-conditioned temporal model. Faceformer \cite{fan2022faceformer} encodes long-term audio context based on transformer \cite{vaswani2017attention} to achieve temporally stable and accurate lip synchronization. CodeTalker \cite{xing2023codetalker} utilizes a codebook with discrete primitives \cite{van2017neural} to reduce the audio-visual mapping uncertainty. Although prior works can synthesize high-fidelity face mesh from speech, they only enable deterministic one-to-one mapping between the speech-to-face mesh. Moreover, they require an identity template mesh as an input to precisely synthesize corresponding facial motions. In this paper, we formulate the task as one-to-many and aim to capture variations in facial motion and jointly synthesize identity and head pose from speech. 

\subsection{2.2. 3D Face Dataset}
3D face dataset \cite{ zhu2021facescape, paysan20093d, cao2013facewarehouse, savran2008bosphorus, yin20063d, zhang2016multimodal, zhang2014bp4d,zhang2013high,fanelli20103,alashkar20143d} is a collection of 3D scans of human faces capturing dynamic facial expressions and poses for applications such as facial recognition, animation, and expression analysis. Yet, only a few datasets \cite{cudeiro2019capture, eth_biwi_00760, cudeiro2019capture,karras2017audio, zhang20193d, richard2021audio} provide synchronized audio, and they are critically scale-limited. First, they do not support a wide range of identities and phonemes. BIWI \cite{eth_biwi_00760} and VOCASET \cite{cudeiro2019capture} are common benchmarks but BIWI \cite{eth_biwi_00760} records only 40 sentences for each of 14 subjects while VOCASET records 255 sentences from 12 subjects. S3DFM \cite{zhang20193d} contains 100 subjects with only one unique utterance for each subject. Multiface \cite{wuu2022multiface} records 36 hours of utterance from 250 subjects but a small fraction of the dataset, comprising 13 subjects speaking 50 sentences, has been made publicly available. Additionally, the recordings are captured in a controlled environment at a studio, with subjects either in static poses or with static expressions. On the other hand, large-scale 2D talking face datasets \cite{afouras2018deep, afouras2018lrs3, chung2018voxceleb2, chung2017lip, nagrani2017voxceleb, wang2020mead, zhang2021flow} are readily accessible and have allowed for the rapid development of speech-driven 2D talking face generation \cite{park2022synctalkface, zhou2021pose, prajwal2020lip, wang2021audio2head, guo2021ad, hong2022depth, liang2022expressive, zhou2019talking, wu2021imitating}. To this end, we transform one of the large-scale 2D talking face datasets, HDTF \cite{zhang2021flow} into a 3D face mesh dataset, namely 3D-HDTF. While the reconstructed 3D faces may not match the fidelity of actual 3D scans, they not only capture a wide variety of facial shapes and motions but also natural head pose and expression while talking. Leveraging the 3D-HDTF, we build a scalable and generalizable speech-driven 3D face animation model.

\subsection{2.3. Diffusion Generation}
Diffusion model is a generative model based on the stochastic diffusion process to generate samples from a probability distribution \cite{sohl2015deep, ho2020denoising}. It starts with a Gaussian noise distribution and iteratively applies a sequence of diffusion steps. Each diffusion step adds Gaussian noise with increasing variance, and the reverse diffusion step removes the Gaussian noise from the sample to get a sample from the target distribution. By iteratively sampling the data distribution starting from the initial noise, it can transform the noise into a complex distribution that resembles the target distribution. With its powerful capability of generating diverse and high-fidelity samples, it has greatly succeeded in image synthesis \cite{dhariwal2021diffusion, song2020denoising, ramesh2022hierarchical, saharia2022photorealistic, rombach2022high, nichol2021glide}. 
Recently, the diffusion model has been applied to talking face generation \cite{stypulkowski2023diffused, shen2023difftalk, du2023dae}
which synthesizes 2D face video conditioned on speech. However, they diffuse at the pixel level to generate high-fidelity images without considering the temporality of the facial motion. Our work is more aligned with motion diffusion models \cite{zhang2022motiondiffuse, ma2022pretrained, tevet2022human} which synthesize human body motion conditioned on textual description. In this paper, we aim to synthesize 3D face mesh conditioned on speech. We take the first approach to employ diffusion to enable precise and one-to-many mapping between speech and face mesh in terms of identity, pose, and motion.

\section{3. Method}
An overview of our method is illustrated in Figure 1. We extract identity representation $\bm{x}_\text{id} \in \mathbb{R}^{3V}$, facial motion representation $\bm{x}_\text{motion}\in \mathbb{R}^{N\times3V}$, and head pose representation $\bm{x}_\text{pose}\in \mathbb{R}^{N\times3}$ from a 3D face mesh sequence $\bm{v} \in \mathbb{R}^{N\times3V}$ of length $N$ and channel size of $V$ vertices. The aim is to learn the joint data distribution of the three representations conditioned on input audio $\bm{a}$, $p(\bm{X}|\bm{a})$ where $\bm{X} = [\bm{x}_\text{id}, \bm{x}_\text{motion}, \bm{x}_\text{pose}]$ is denoted as the face representations. We introduce 3D face mesh diffusion model which jointly diffuses the face representations and enables one-to-many 3D face animation while aligning the lip motion with the input speech. At inference, the face representations are sampled from the learned data distribution, where each component of the face representations can be individually controlled.

\begin{figure*}[t!]
	\begin{minipage}[b]{\linewidth}
		\centering		
            \centerline{\includegraphics[width=15cm]{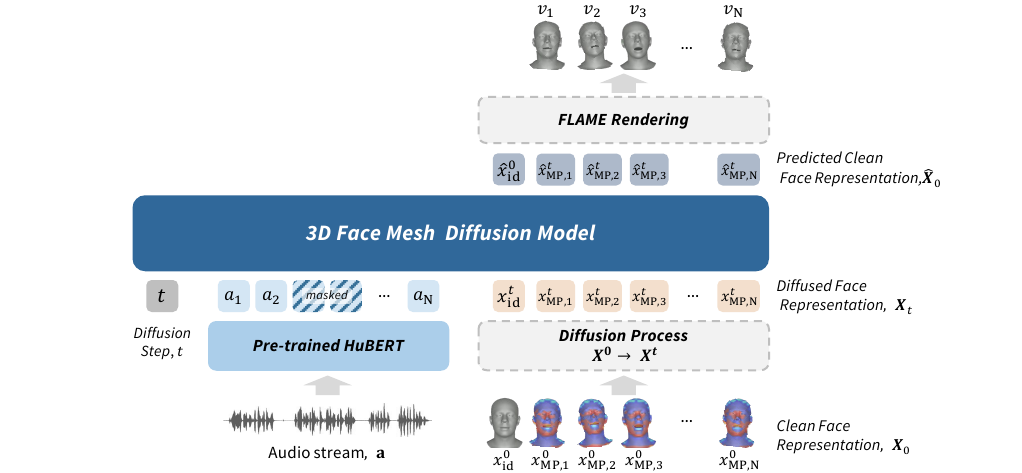}}
	\end{minipage}
	\vspace{-0.3cm}
	\caption{Overview of our DF-3DFace framework. The clean face representation $\bm{X}_{0}=[\bm{x}_\text{id}^0, \bm{x}_\text{MP, 1:N}^0]$ is noised into $\bm{X}_{t}$ through the diffusion process. Given the diffusion timestep $t$, randomly masked HuBERT audio representation $\bm{a}$, and diffused face representation $\bm{X}_{t}$, 3D face mesh diffusion model learns to predict clean face representation $\hat{\bm{X}}_{0}$. The predicted clean face representation $\hat{\bm{X}}_{0}$ is then rendered as a sequence of 3D face mesh. Note that the motion pose representations $\bm{x}_\text{MP, 1:N}$ are visualized as heat maps where regions with higher values are in the color red while the lower are in the color blue.}
	\label{fig:1}
  \vspace{-0.15cm}
\end{figure*}

\subsection{3.1. 3D Face Mesh Representations}
To synthesize identity and pose along with facial motion, we disentangle a 3D face mesh sequence $\bm{v} \in \mathbb{R}^{N\times3V}$ into identity representation $\bm{x}_\text{id}$, pose representation $\bm{x}_\text{pose}$, and facial motion representation $\bm{x}_\text{motion}$. In this section, we explain how we have factored out each of the components in 3D mesh space so that the face mesh can be re-rendered without losing information. 

\subsubsection{Identity Representation}
Identity contains facial features and structures that are unique to each individual. To eliminate the variance of the mesh due to the change in head pose, we first transform the face mesh sequence $\bm{v}$ into zero head pose by applying the Linear blend skinning (LBS) \cite{loper2015smpl}. By forcing the global pose vector of the reconstructed pose parameter $\bm{\theta}$ to zero, the face mesh sequence $\bm{v}$ is transformed into zero pitch, roll, and yaw. 
We take the average of randomly sampled $K$ zero pose face meshes $\bm{x}_\text{zeropose}$ for individual subjects as the identity representation, $ \bm{x}_\text{id} = \frac{1}{K}\sum_{i}^{K}{x}_\text{zeropose}^{i}.$

\subsubsection{Pose Representation} Pose refers to the orientation of the face in 3D space. From the 3D face reconstruction method \cite{DECA:Siggraph2021}, we obtain the pose parameter $\bm{\theta}$ for each timestep $t$, from which we extract the global rotation vector, $\bm{x}_\text{pose} = \{{\theta}_\text{global}^{t}\}_{t=1}^{N}.$



\subsubsection{Motion Representation} Facial motion, which we shorten as motion, includes dynamics of the inner face including facial expressions reflecting emotion, movement of the mouth and jaw in sync with speech, eyebrow raising, and eye blinking.
To capture such dynamics, we represent the motion as the deviation of the vertices in the zero pose face mesh ${x}_\text{zeropose}^{t}$ from the corresponding identity representation $\bm{x}_\text{id}$ as $\bm{x}_\text{motion} = \{ x_\text{zeropose}^{t} - \bm{x}_\text{id}\}_{t=1}^{N}.$

\subsection{3.2. 3D Face Mesh Diffusion} 
To enable one-to-many mapping from speech $\bm{a}$ to the face representations $\bm{X}=[\bm{x}_\text{id}, \bm{x}_\text{motion}, \bm{x}_\text{pose}]$ as obtained in section 3.1, we model the data distribution $p(\bm{X}|\bm{a})$ based on diffusion \cite{ho2020denoising}. 

The diffusion defines a forward noising process $q(x_t|x_{t-1})$, for $t \in [0,T]$ as a Gaussian distribution following a Markov chain, 
\begin{equation}
{
q(x_{t} | x_{t-1}) = \mathcal{N}(\sqrt{\alpha_{t}}x_{t-1},(1-\alpha_{t})I),
}
\end{equation}
where clean sample $x_{0}$ is repeatly corrupted into pure noise $x_{T}=\mathcal{N}(0, I)$ controlled by a predefined variance scheduler $\alpha_{t}$. 
The denoising step, which is the reverse of the noising step, is also defined as a Gaussian distribution, 
\begin{equation}
{
p(x_{t-1} | x_{t}) = \mathcal{N}(\mu_{\theta}(x_{t}, t), \sigma_{\theta}(x_t, t)I),
}
\end{equation}
where the pure noise $x_T$ is iteratively denoised back to a clean sample $x_0$ based on the mean $\mu_{\theta}(\cdot)$ and variance $\sigma_{\theta}(\cdot)$ parameterized using a deep neural network. 

In our context, we model audio-conditioned face representation $p(\bm{X}|\bm{a})$ with the reverse diffusion process of predicting clean face representation $\bm{X}_{0}$ from noised face representations $\bm{X}_{t}$ conditioned on audio $\bm{a}$ ($x_{0}-\text{formulation}$ \cite{ramesh2022hierarchical}). During training, we uniformly sample $t \in [0, T]$ and obtain noised face representation $\bm{X}_{t}$ at diffusion timestep $t$ from $q(\bm{X}_{t} | \bm{X}_{0}) = \mathcal{N}(\sqrt{\bar{\alpha}_{t}}\bm{X}_{0},(1-\bar{\alpha}_{t})I)$ where $\bar{\alpha}_{t}=\prod^{T}_{t=1}{\alpha}_{t}$. 
Given the diffusion step $t$, audio $\bm{a}$, and diffused face representation $\bm{X}_{t}$, we implement a diffusion model $G_{\theta}$, based on transformer-encoder \cite{vaswani2017attention}, to predict the clean face representation  $\hat{\bm{X}}_{0}$ as follows, 
\begin{equation}
\hat{\bm{X}}_0 = G_{\theta}(t, \bm{a}, \bm{X}_t).  
\end{equation}
The audio representation $\bm{a} \in \mathbb{R}^{N_{a}\times Z_{a}}$ is extracted using the state-of-the-art ASR model, Hu-BERT \cite{hsu2021hubert}. The face representation $\bm{X} \in \mathbb{R}^{(N+1)\times(3V+3)}$ is formulated as a concatenation of the motion representation $\bm{x}_\text{motion} \in \mathbb{R}^{N \times 3V}$ and pose representation $\bm{x}_\text{pose} \in \mathbb{R}^{N \times 3}$ along channel axis, denoted as the motion pose represention $\bm{x}_\text{MP} \in \mathbb{R}^{N\times(3V+3)}$,  followed by a concatenation of the identity representation $\bm{x}_\text{id} \in \mathbb{R}^{3V}$ along the time axis (with additional zero-padded 3 dimension).  
For each frame in the diffused face representation $\bm{X}_{t}$, we individually project identity, motion, and pose representations to match the diffusion latent space $\mathbb{R}^{C}$. The diffusion timestep $t$ and audio representation $\bm{a}$ are also individually projected and $\bm{a}$ is resampled to match the length of the mesh sequence $N$. The projected audio representation and face representation are summed with their respective modality embeddings and positional embeddings. Then, they are concatenated along the time axis with the projected diffusion timestep $t$ as the first token, forming the input tokens $\mathbb{R}^{(2N+2)\times C}$. From the output, we exclude the first $(N+1)$ output tokens (corresponding to $t$ and $\bm{a}$) and project the rest back to the original mesh space to obtain the predicted clean face representations $\hat{\bm{X}}_0 \in \mathbb{R}^{(N+1)\times(3V+3)}$. 

We employ the standard denoising diffusion loss,
\begin{equation}
{
\mathcal{L}_\text{face} = \mathbb{E}_{(\bm{X}_0, \bm{a}), t \sim [1,T]}\Bigr[\| \bm{X}_0 - G_{\theta}(t, \bm{a}, \bm{X}_t)\|_2^2\Bigr].
}
\end{equation}

\begin{table*}
	\renewcommand{\arraystretch}{1}
	\renewcommand{\tabcolsep}{1.2mm}
\centering
\small
\centering
\resizebox{0.8\linewidth}{!}{
\begin{tabular}{lcccc}
\toprule
\makecell{\multirow{2}{*}{Method}} &  
\makecell{\multirow{2}{*}{Dataset}}  & 
\multicolumn{1}{p{3cm}}{\centering Avg Lip Vertex Error$\downarrow$ \\ ($\times 10^{-5}\text{mm}$) } & 
\multicolumn{1}{p{3cm}}{\centering Max Lip Vertex Error$\downarrow$ \\ ($\times 10^{-5}\text{mm}$)} & 
\multicolumn{1}{p{3cm}}{\centering Non-Lip Deviation$\downarrow$ \\ ($\times 10^{-4}\text{mm}$)}  \\
\midrule 
{MeshTalk \cite{richard2021meshtalk}}& VOCASET  &  2.734 & 6.910 & 1.818  \\
{Faceformer \cite{fan2022faceformer}}& VOCASET & 2.655 &  7.521 & {1.790}\\
{CodeTalker \cite{xing2023codetalker}}&  VOCASET & {2.492} & \bf{6.406} & \bf{1.749} \\
{DF-3DFace}& VOCASET  &  \bf{2.489} & {6.545} & 2.253  \\
\midrule  
{MeshTalk \cite{richard2021meshtalk}} & 3D-HDTF$\ast$  &  0.646 & 1.567 & 3.295  \\
{Faceformer \cite{fan2022faceformer}} & 3D-HDTF$\ast$  &  {0.510} & {1.214} & {2.823}  \\
{CodeTalker \cite{xing2023codetalker}} & 3D-HDTF$\ast$  &  2.679 & 5.617 & 4.449  \\
{DF-3DFace} &  3D-HDTF$\ast$ & \bf{0.430}  & \bf{1.037} & \bf{0.970}   \\
\midrule 
{DF-3DFace w/o $\bm{x}_\text{id}$} &  3D-HDTF  & 0.421 & 1.029 & 0.635   \\
{DF-3DFace w/o $\bm{x}_\text{pose}$} &  3D-HDTF  & 0.423 & 1.045 & 0.644   \\
{DF-3DFace w/o $\mathcal{L}_\text{lip}$}& 3D-HDTF  & 0.434 & 1.041 & 0.648  \\
{DF-3DFace w/o $\mathcal{L}_\text{sync}$}& 3D-HDTF  & 0.452 & 1.073 & 0.647  \\
{DF-3DFace w/o $\mathcal{L}_\text{pose}$}& 3D-HDTF  &  0.430 & 1.038 & 0.650 \\
{DF-3DFace w/o $\mathcal{M}_\text{cond}$}& 3D-HDTF  &  0.433 & 1.037 & 0.636  \\
{DF-3DFace} &  3D-HDTF  &  0.496 & 1.204 & 0.776  \\
\bottomrule
\end{tabular} }
\vspace{0.0in}
\caption{\label{tb:AMT-BIWI} Quantitative evaluation on VOCASET, 3D-HDTF$\ast$, and 3D-HDTF.}
\vspace{-0.15cm}
\end{table*}

\subsection{3.3. Masked Conditioning} 
We introduce masked conditioning $\mathcal{M}_\text{cond}$ to improve diversity while minimizing trade-offs in the fidelity of lip synchronization. In practice, classifier-free guidance \cite{ho2022classifier} learns both conditioned and unconditioned distributions by randomly setting the condition $c=\emptyset$ for 10\% of the samples during training and sampling the interpolation of the two variants during inference. It has been used in tasks that have more freedom in generation such as class-conditioned and text-conditioned image generation \cite{saharia2022photorealistic, nichol2021glide, dhariwal2021diffusion}, video generation \cite{ho2022imagen, ho2022video} and human body motion generation \cite{zhang2023remodiffuse, ma2022pretrained}. 
However, our task is more tightly bound by the condition as the mesh has to timely align with the audio. 
Simply setting the audio to null would hurt the lip sync performance. Thus, we apply random mask $m$ to 10\% of the audio representations $\bm{a}$ during training and interpolate between the masked conditioned and unmasked conditioned distributions with the hyperparameter $s$ during inference as follows, 
\begin{align}
G_{\theta}(t, \bm{a}, &\bm{X}_{t}) = G_{\theta}(t, m(\bm{a}), \bm{X}_{t}) \\&+ s \cdot (G_{\theta}(t, \bm{a}, \bm{X}_{t}) - G_{\theta}(t, m(\bm{a}), \bm{X}_{t})).
\end{align}

\subsection{3.4. Mesh and Audio Synchronization} 
To accurately synchronize the 3D face mesh with the audio, we propose a 3D mesh-based sync expert.
We train a sync expert $\mathcal{S}_\text{sync}$ such that it measures the likelihood that the input audio and the 3D face mesh is in-sync, inspired by \cite{prajwal2020lip}. Specifically, we randomly select segments of face representation ${x}_\text{motion}^{i:i+n}$ of length $n$ and apply a lip mask $m_\text{lip}$ which selects vertices pertaining to the mouth region in the motion representation $\bm{x}_\text{motion}$. The sync expert calculates the cosine similarity distance between the mouth motion and audio representations $a^{i:i+n}$ which is extracted from HuBERT \cite{hsu2021hubert}. The sync expert is trained to minimize the distance between synchronized pairs while maximizing the distance between out-of-sync pairs. More training details can be found in the supplementary. The trained sync expert is then fixed and used to guide the DF-3DFace to enhance speech-lip synchronization of the synthesized facial motions $\hat{{x}}_\text{motion}^{i:i+n}$ with the following sync loss $\mathcal{L}_\text{sync}$, 
\begin{equation}
{
\mathcal{L}_\text{sync} = \mathbb{E}_{(\hat{\bm{x}}_\text{motion}, \bm{a}),i \sim [1,N-n]}[\mathcal{S}_\text{sync}(m_\text{lip} \cdot \hat{{x}}_\text{motion}^{i:i+n}, a^{i:i+n})].
}
\end{equation}

\subsection{3.5. Auxiliary Loss} 
To place greater emphasis on the mouth and pose, which constitute the majority of the dynamics in the mesh, we additionally apply lip loss $\mathcal{L}_\text{lip}$, and pose loss $\mathcal{L}_\text{pose}$ as follows,  
\begin{equation}
{
\mathcal{L}_\text{lip} = \mathbb{E}_{\hat{\bm{x}}_\text{motion}}[\| m_\text{lip} \cdot (\bm{x}_\text{motion} -\hat{\bm{x}}_\text{motion}) \|_2^2],
}
\end{equation}
\vspace{0.5pt}
\begin{equation}
{
\mathcal{L}_\text{pose} = \mathbb{E}_{\hat{\bm{x}}_\text{pose}, i \sim [1,N-1]}[\| (\ {x}_\text{pose}^{i+1} - {x}_\text{pose}^{i}) - (\hat{x}_\text{pose}^{i+1} - \hat{x}_\text{pose}^{i})\|_2^2]. 
}
\end{equation}
The lip loss $\mathcal{L}_\text{lip}$ measures the MSE loss between the predicted and ground facial motion ($\hat{\bm{x}}_\text{motion}$ and $\bm{x}_\text{motion}$ respectively) of the mouth region using the lip mask $m_\text{lip}$. 
As the movement of the pose is more dynamic than the facial motion, the pose transition needs to be smooth so we apply a pose loss $\mathcal{L}_\text{pose}$ which brings the difference of adjacent pose representations of the ground truth mesh close to that of the generated mesh. Note that we have dropped the notation for the diffusion timestep of $0$. 
The losses are scaled with respective hyperparameters $\lambda$ and summed to give the final training loss, 
\begin{equation}
{
\mathcal{L} = \lambda_\text{face}\mathcal{L}_\text{face} + 
\lambda_\text{sync}\mathcal{L}_\text{sync} + \lambda_\text{lip}\mathcal{L}_\text{lip} + \lambda_\text{pose}\mathcal{L}_\text{pose}.
}
\end{equation}

\begin{figure*}[t!]
	\begin{minipage}[b]{\linewidth}
		\centering		
            \centerline{\includegraphics[width=14cm]{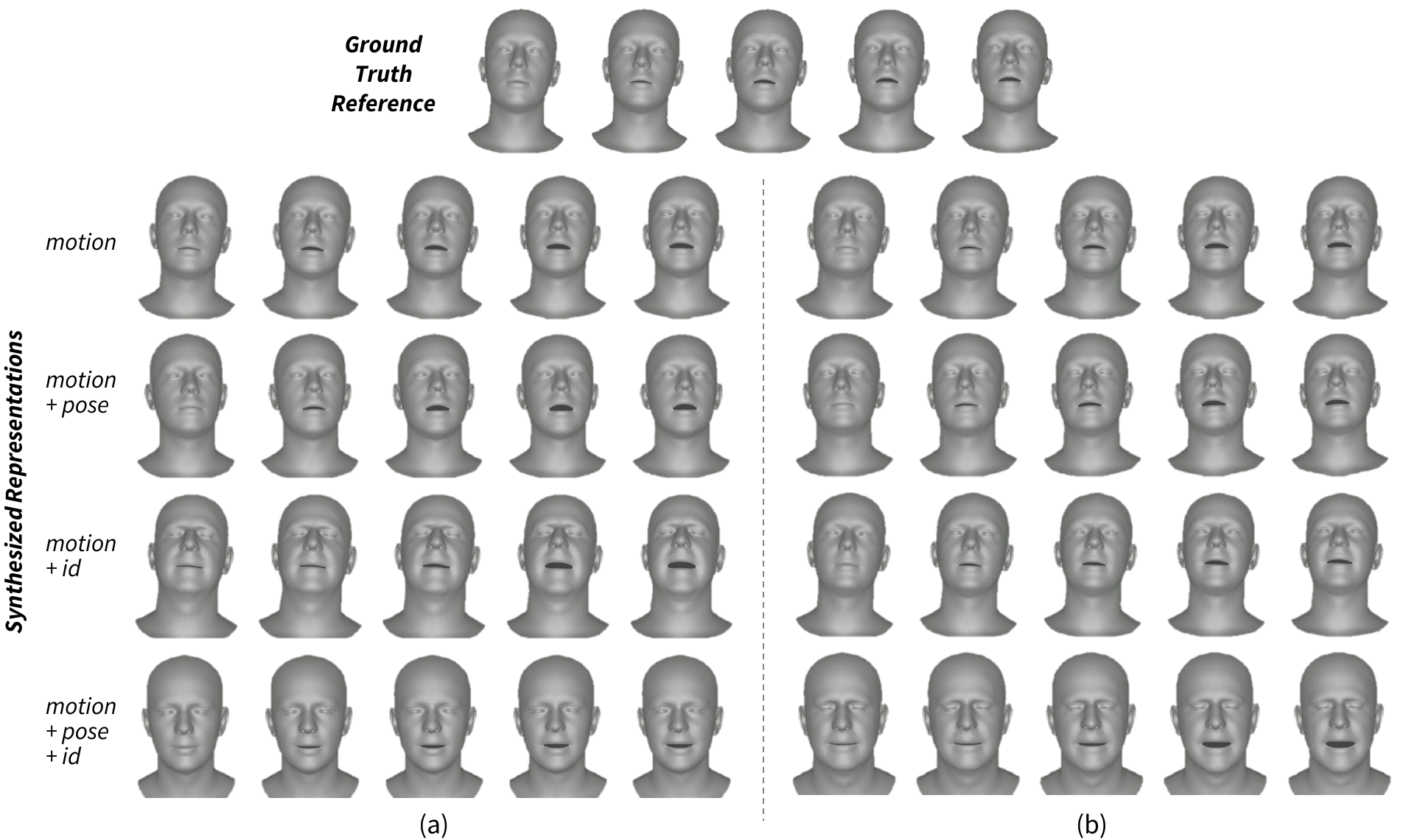}}
	\end{minipage}
	\vspace{-0.5cm}
	\caption{Generation results by DF-3DFace where different components of the face representations have been synthesized as indicated on the left, and sampling has been performed twice in (a) and (b). While consistently achieving accurate lip synchronization, DF-3DFace generates diverse 3D face animations with controllable identity, motion, and pose. We encourage readers to zoom in for details.}
	\label{fig:1}
 \vspace{-0.1cm}

\end{figure*}

\begin{table*}
	\renewcommand{\arraystretch}{1.0}
	\renewcommand{\tabcolsep}{1.2mm}
\centering
\resizebox{0.8\linewidth}{!}{
\begin{tabular}{lccccccc}
\toprule
\makecell{\multirow{2}{*}{{Comparison}}} & \multicolumn{2}{c}{\centering {VOCASET}} & \multicolumn{2}{c}{\centering {3D-HDTF$\ast$}} & \multicolumn{2}{c}{\centering {3D-HDTF}} \\ \cmidrule(lr){2-3} \cmidrule(lr){4-5}  \cmidrule(lr){6-7}  
 &
\multicolumn{1}{p{2cm}}{\centering Realism} & 
\multicolumn{1}{p{2cm}}{\centering Lip Sync} & 
\multicolumn{1}{p{2cm}}{\centering Realism} & 
\multicolumn{1}{p{2cm}}{\centering Lip Sync} &
\multicolumn{1}{p{2cm}}{\centering Realism} & 
\multicolumn{1}{p{2cm}}{\centering Lip Sync} \\
\midrule 
{\centering DF-3DFace vs GT}&  $40.28\pm3.66$  & $57.78\pm4.20$ & $50.72\pm1.93$  & $51.71\pm2.13$ & $65.56\pm2.45$ & $69.39\pm1.99$ \\
\midrule 
{\centering DF-3DFace vs MeshTalk}& $70.56\pm1.87$ & $68.89\pm1.97$ & $72.58\pm1.92$ & $73.29\pm1.63$ & -- & -- \\
{\centering DF-3DFace vs Facformer}& $48.56\pm2.96$  & $51.11\pm2.13$ & $60.47\pm2.92$  & $62.39\pm2.29$ & -- & --  \\
{\centering DF-3DFace vs CodeTalker}& $49.17\pm2.18$  & $50.89\pm2.21$ & $63.69\pm1.89$  & $64.10\pm1.88$ & -- & -- \\
\bottomrule 
\end{tabular}  }
\vspace{-0.0in}
\caption{\label{tb:AMT-BIWI} User study on VOCASET, 3D-HDTF$\ast$, and 3D-HDTF. The score indicates preference of DF-3DFace over others in \%}
\vspace{-0.15cm}
\end{table*}

\subsection{3.6. Sampling} 
Starting from a Gaussian noise $\bm{X}_{T} \sim \mathcal{N}(0, I)$, we iteratively sample $\bm{X}_{t-1} \sim p_{\theta}(\bm{X}_{t-1}|\bm{X}_{t}, \bm{a})$ until the final clean face representation $\hat{\bm{X}}_{0}$ is reached. $p_{\theta}(\bm{X}_{t-1}|\bm{X}_{t},\bm{a})$ follows a Gaussian distribution $\mathcal{N}(\mu_{\theta},\sigma I)$, where $\mu_{\theta}=\sqrt{\bar{\alpha}_{t-1}}G_{\theta}(t, \bm{a}, \bm{X}_t)$ and $\sigma=(1-\bar{\alpha}_{t-1})I$ are the mean and variance, induced from the predicted clean face representation $\hat{\bm{X}}_{0}=G_{\theta}(t, \bm{a}, \bm{X}_t)$ at diffusion timestep $t$. 
Starting from the pure noise and iteratively sampling from the predicted Gaussian distributions leads to multiple different facial animations across multiple sampling processes. 
Although the face representation as a whole is jointly diffused, we enable individual control of the identity, pose, and motion during sampling. Specifically, instead of generating all three representations given an audio input $\bm{a}$, we can give any combinations of the identity, pose, and motion as target references. For example, to generate a mesh of a reference identity speaking the input audio, we input the target reference identity instead of sampling from the Gaussian noise, and at each iteration during sampling, we overwrite the identity representations of the output with its target reference. 
Therefore, DF-3DFace can not only synthesize diverse identities, pose, and motions from a single speech but also synthesize with the given reference. This is in contrast to the existing approach where the scheme is limited to animating the given identity in static head pose. 

\subsection{3.7. Mesh Rendering} 
Once we obtain the face representation $\bm{X}=[\bm{x}_\text{id}, \bm{x}_\text{motion}, \bm{x}_\text{pose}]\in \mathbb{R}^{(N+1)\times(3V+3)}$ from the diffusion model $G_{\theta}$, we render it back to the original face mesh $\bm{v}\in \mathbb{R}^{N\times3V}$ by reversing the steps in section 3.1. We first add identity representation $\bm{x}_\text{id}$ to each of the motion representation $\bm{x}_\text{motion}$ to get zero pose face mesh $\bm{x}_\text{zeropose}$. Then we deform the zero pose face mesh $\bm{x}_\text{zeropose}$ with the pose representation $\bm{x}_\text{pose}$ based on LBS \cite{loper2015smpl}.


\section{4. Experiments}
\subsection{4.1. Datasets}
\subsubsection{3D-HDTF} 
Since existing 3D face databases are critically limited in size and scope to hold a natural and diverse range of 3D faces while talking, we construct a 3D-HDTF. 
3D-HDTF is synthesized from a large in-the-wild high-resolution talking face audio-visual dataset (HDTF) \cite{zhang2021flow}. It contains 300 speakers with 15.8 hours of approximately 10,000 utterances, captured at 30 fps and collected from youtube. We use the 3D face reconstruction method \cite{DECA:Siggraph2021} to obtain mesh in FLAME topology as has been applied in \cite{cudeiro2019capture, DECA:Siggraph2021, sanyal2019learning, ghosh2020gif, cao2005expressive, ranjan2018generating}. Please refer to the supplementary for the implementation detail of the reconstruction model. 

\subsubsection{3D-HDTF$\ast$} 
We additionally introduce a zero-pose version of the 3D-HDTF namely 3D-HDTF* to fairly compare with the prior works that can not synthesize the head pose. Each mesh is transformed into zero head pose position using the linear blend skinning \cite{loper2015smpl}. 

\subsubsection{VOCASET} VOCASET \cite{cudeiro2019capture} is a popular 3D face scan dataset. It has 12 speakers with 480 3D facial motion sequences captured at 60fps with a duration of 3 to 4 seconds each. There are 255 unique sentences, some of which are shared between the speakers. We use the train, validation, and test splits provided by VOCASET. 

\begin{table*}
	\renewcommand{\arraystretch}{1.0}
	\renewcommand{\tabcolsep}{1.2mm}
\centering
\resizebox{0.7\linewidth}{!}{
\begin{tabular}{lllcccS[table-format=2.3]}
\toprule
\makecell{\multirow{2}{*}{Method}} & 
\makecell{\multirow{2}{*}{Input}} & 
\makecell{\multirow{2}{*}{Output}} & 
\multicolumn{1}{p{2cm}}{\centering $\text{Mult}_\text{id}$$\uparrow$\\ {($\times 10^{-6}\text{mm}$)}} & 
\multicolumn{1}{p{2cm}}{\centering $\text{Mult}_\text{motion}$$\uparrow$\\ {($\times 10^{-7}\text{mm}$)}} & 
\multicolumn{1}{p{2cm}}{\centering $\text{Mult}_\text{pose}$$\uparrow$\\ {($\times 10^{-3}\text{mm}$)}} &
\multicolumn{1}{p{2cm}}{\centering $\text{Mult}_\text{mesh}$$\uparrow$\\ {($\times 10^{-6}\text{mm}$)}} \\
\midrule 
\multirow{4}{*}{\makecell{DF-3DFace}} &  {aud, id, pose}  &  {motion}  &  0.000  & 1.813 & 0.000 & 0.197 \\
& {aud, pose} &  {motion, id}  &  1.834   & 1.916 & 0.000 & 2.091 \\
&  {aud, id}  &  {motion, pose}  &  0.000 & 1.813 & 2.910 & 80.779 \\
& {aud} & {motion, pose, id}   & 1.834   & 1.916 & 5.212 &  152.130 \\
\midrule 
{+ w/o $\mathcal{M}_\text{cond}$}&  {aud} & {motion, pose, id}  &  0.313 & 0.660 & 1.311 & 31.601 \\
\bottomrule
\end{tabular}}
\vspace{0.0in}
\caption{\label{tb:AMT-BIWI} Multimodality Evaluation on 3D-HDTF.}
\vspace{-0.15cm}
\end{table*}

\subsection{4.2. Evaluation Metrics}
\subsubsection{Lip Synchronization} We follow the evaluation protocol employed in state-of-the-works \cite{cudeiro2019capture, richard2021meshtalk, fan2022faceformer, xing2023codetalker} to measure lip synchronization. The lip synchronization is measured with lip-vertex error which is the maximal L2 distance of lip vertices between the prediction and the ground truth. 

\subsubsection{Non-Lip Dynamics Deviation} We implement non-lip dynamics deviation (NLDD) motivated by \cite{xing2023codetalker} to evaluate the dynamics of the non-lip part of the face mesh. It computes the difference in the standard deviation of element-wise L2 norm along the temporal axis between the predicted and ground truth non-lip vertices. 

\subsubsection{Multimodality} We evaluate the variability of the generation by calculating how much the generation diversifies within each given audio as proposed in  \cite{guo2020action2motion}. Given an input audio, it generates two subsets with the same sample size of 100 and calculates the L2 distance between the two subsets. We measure the multimodality in terms of mesh $\bm{v}$, identity $\bm{x}_\text{id}$, motion $\bm{x}_\text{motion}$, and pose $\bm{x}_\text{pose}$. 

\subsection{4.3. Quantitative Evaluation}
\subsubsection{Lip Synchronization \& Non-lip Deviation}
We quantitatively compare our DF-3DFace with state-of-the-art methods, Meshtalk \cite{richard2021meshtalk}, Faceformer \cite{fan2022faceformer}, and Codetalker \cite{xing2023codetalker} in Table 1. Implementation details of DF-3DFace and the comparison methods are provided in the supplementary. Since these methods only synthesize facial motion given the identity and audio, we evaluate the facial motion on the zero-pose datasets (VOCASET and 3D-HDTF$\ast$). On both datasets, DF-3DFace achieves lower lip vertex error, producing more synchronized lip movement than the state-of-the-art methods. As our method jointly synthesizes the identity, the non-lip deviation is apparently higher on VOCASET which contains minimal variance in the non-lip region. But it achieves significantly superior performance on the 3D-HDTF$\ast$ which encompasses more variations in the facial shapes and motions. The result demonstrates the effectiveness of our method in generating a wide range of 3D face meshes from speech.

\subsubsection{Variability \& Controllability}
We validate the variability and controllability in Table 3. DF-3DFace introduces stochasticity through the diffusion process and enables one-to-many mapping as demonstrated by the multimodality scores. This is in contrast to prior works that learn the deterministic one-to-one mapping between speech and mesh. We also measure the multimodality of each component of the face representations when provided with different references at the input. The result indicates that giving identity as a reference constrains the variance of the motion and pose while giving pose as a reference does not affect the variance of the motion and identity. This is because the range of possible motion and pose can be narrowed down with the specified identity. The mesh multimodality is the highest when only audio is given and decreases as more references are incorporated. Also, the result verifies independent controllability of the identity, motion, and pose because those provided as the references have zero multimodality, allowing variance only in the non-referenced components. 

\subsection{4.4. Qualitative Evaluation}
We refer to the demo video for detailed qualitative evaluation as the sync quality and naturalness are best viewed over time with the audio. Compared with the baseline methods, our method not only generates the most synchronized lip movements but also captures the natural and diverse range of facial motions present in the 3D-HDTF$\ast$ dataset. On the 3D-HDTF$\ast$, Faceformer and Meshtalk portray motions confined to the lip region only, which makes the animation unrealistic. CodeTalker captures the motions across the face including the eyeblinks but sometimes the face is deformed arbitrarily. In contrast, our method portrays the dynamics over the entire face most similar to the ground truth, based on the diffusion process which accurately models the complex relationship between speech and 3D face mesh from the dataset. 

Generation results in Figure 2 further illustrate variability and controllability of identity, pose, and motion. We synthesize with different references from the ground truth, and sample twice in Figure 2 (a) and (b). The second row is when identity and pose are provided as references and only motion is synthesized. The pose and identity align with the ground truth reference while there is variance only in motion where (a) and (b) have different eye blinks. When the pose is synthesized in the third row, it exhibits variations in the pose where a) is posing right upward and (b) straight upward. When identity is synthesized, there are subtle variations in the shape of the skull and the size of the jawline as shown in the last two rows. While capturing variations in facial attributes from a single speech and allowing controllability, DF-3DFace achieves synchronized lip. Please refer to the demo video and supplementary for detailed demonstrations. 

\subsubsection{User Study}
We conduct A/B tests for perceptual evaluation in Table 2. We asked 20 participants to choose the better sample in terms of realism and lip synchronization when presented with side-by-side clips of our approach versus others. We randomly select 15 to 20 samples for each dataset, giving a total of 180 clips. Note that we only evaluate the 3D-HDTF against the ground truth because other methods do not support pose synthesis. Our method surpasses lip synchronization, and the performance gap is even larger when synthesizing more dynamic animation in 3D-HDTF. Interestingly, our method even outperforms the ground truth, indicating that it models the data distribution with high-fidelity lip synchronization based on the mesh-audio synchronization learning, compensating for the reconstruction error that occurred during the data construction.

\subsection{4.5. Ablation Study} 
\subsubsection{Identity and Pose Representations}
Please note that the following ablations in Table 1 are conducted without masked conditioning $\mathcal{M}_\text{cond}$. Without learning $\bm{x}_\text{id}$ or $\bm{x}_\text{pose}$, lip synchronization and non-lip deviation both improve as the model can focus on learning the facial motion. Yet, it enables synthesis and manipulation over identity and pose with negligible performance overhead. 

\subsubsection{Loss Functions}
Both lip loss $\mathcal{L}_\text{lip}$ and sync loss $\mathcal{L}_\text{sync}$ contribute to lip synchronization and the effect of the sync loss is greater than the lip loss. The pose loss  $\mathcal{L}_\text{pose}$ improves non-lip deviation with a trade-off between lip synchronization. The three losses altogether give the best performance. 

\subsubsection{Masked Conditioning}
Masked conditioning effectively boosts diversity where mesh multimodality $\text{Mult}_\text{mesh}$ increases by nearly 5 times as shown in Table 3 with a tradeoff in the lip vertex error as shown in Table 1. Note that the experiments were performed with $s=1$. Please refer to the supplementary for comprehensive experiments of the diversity hyperparameter $s$ and \% of masking. 



\section{5. Conclusion}
We present DF-3DFace, a novel framework for one-to-many speech-synchronized 3D facial animation. It enables variations of the 3D face from a single speech while ensuring lip synchronization by exploiting audio-mesh synchronization and masked conditioning. Our diffusion-based approach effectively captures the natural and diverse range of 3D faces present in the constructed 3D-HDTF. Also, DF-3DFace models the identity, pose, and motion for a comprehensive animation and additionally allows controllability to each component. Through extensive experiments, we validate the effectiveness of the proposed DF-3DFace in incorporating diversity and achieving high-fidelity 3D facial animation.

\bigskip

\bibliography{aaai24}

\begin{thebibliography}{71}
\providecommand{\natexlab}[1]{#1}

\bibitem[{Afouras et~al.(2018)Afouras, Chung, Senior, Vinyals, and
  Zisserman}]{afouras2018deep}
Afouras, T.; Chung, J.~S.; Senior, A.; Vinyals, O.; and Zisserman, A. 2018.
\newblock Deep audio-visual speech recognition.
\newblock \emph{IEEE transactions on pattern analysis and machine
  intelligence}, 44(12): 8717--8727.

\bibitem[{Afouras, Chung, and Zisserman(2018)}]{afouras2018lrs3}
Afouras, T.; Chung, J.~S.; and Zisserman, A. 2018.
\newblock LRS3-TED: a large-scale dataset for visual speech recognition.
\newblock \emph{arXiv preprint arXiv:1809.00496}.

\bibitem[{Alashkar et~al.(2014)Alashkar, Amor, Daoudi, and
  Berretti}]{alashkar20143d}
Alashkar, T.; Amor, B.~B.; Daoudi, M.; and Berretti, S. 2014.
\newblock A 3D dynamic database for unconstrained face recognition.
\newblock In \emph{5th International Conference and Exhibition on 3D Body
  Scanning Technologies}.

\bibitem[{Cao et~al.(2013)Cao, Weng, Zhou, Tong, and
  Zhou}]{cao2013facewarehouse}
Cao, C.; Weng, Y.; Zhou, S.; Tong, Y.; and Zhou, K. 2013.
\newblock Facewarehouse: A 3d facial expression database for visual computing.
\newblock \emph{IEEE Transactions on Visualization and Computer Graphics},
  20(3): 413--425.

\bibitem[{Cao et~al.(2005)Cao, Tien, Faloutsos, and Pighin}]{cao2005expressive}
Cao, Y.; Tien, W.~C.; Faloutsos, P.; and Pighin, F. 2005.
\newblock Expressive speech-driven facial animation.
\newblock \emph{ACM Transactions on Graphics (TOG)}, 24(4): 1283--1302.

\bibitem[{Chung, Nagrani, and Zisserman(2018)}]{chung2018voxceleb2}
Chung, J.~S.; Nagrani, A.; and Zisserman, A. 2018.
\newblock Voxceleb2: Deep speaker recognition.
\newblock \emph{arXiv preprint arXiv:1806.05622}.

\bibitem[{Chung and Zisserman(2017)}]{chung2017lip}
Chung, J.~S.; and Zisserman, A. 2017.
\newblock Lip reading in the wild.
\newblock In \emph{Computer Vision--ACCV 2016: 13th Asian Conference on
  Computer Vision, Taipei, Taiwan, November 20-24, 2016, Revised Selected
  Papers, Part II 13}, 87--103. Springer.

\bibitem[{Cudeiro et~al.(2019)Cudeiro, Bolkart, Laidlaw, Ranjan, and
  Black}]{cudeiro2019capture}
Cudeiro, D.; Bolkart, T.; Laidlaw, C.; Ranjan, A.; and Black, M.~J. 2019.
\newblock Capture, learning, and synthesis of 3D speaking styles.
\newblock In \emph{Proceedings of the IEEE/CVF Conference on Computer Vision
  and Pattern Recognition}, 10101--10111.

\bibitem[{Dhariwal and Nichol(2021)}]{dhariwal2021diffusion}
Dhariwal, P.; and Nichol, A. 2021.
\newblock Diffusion models beat gans on image synthesis.
\newblock \emph{Advances in Neural Information Processing Systems}, 34:
  8780--8794.

\bibitem[{Du et~al.(2023)Du, Chen, He, Tan, Chen, Yu, Zhao, and
  Bian}]{du2023dae}
Du, C.; Chen, Q.; He, T.; Tan, X.; Chen, X.; Yu, K.; Zhao, S.; and Bian, J.
  2023.
\newblock DAE-Talker: High Fidelity Speech-Driven Talking Face Generation with
  Diffusion Autoencoder.
\newblock \emph{arXiv preprint arXiv:2303.17550}.

\bibitem[{Fan et~al.(2022{\natexlab{a}})Fan, Lin, Saito, Wang, and
  Komura}]{fan2022faceformer}
Fan, Y.; Lin, Z.; Saito, J.; Wang, W.; and Komura, T. 2022{\natexlab{a}}.
\newblock Faceformer: Speech-driven 3d facial animation with transformers.
\newblock In \emph{Proceedings of the IEEE/CVF Conference on Computer Vision
  and Pattern Recognition}, 18770--18780.

\bibitem[{Fan et~al.(2022{\natexlab{b}})Fan, Lin, Saito, Wang, and
  Komura}]{fan2022joint}
Fan, Y.; Lin, Z.; Saito, J.; Wang, W.; and Komura, T. 2022{\natexlab{b}}.
\newblock Joint audio-text model for expressive speech-driven 3d facial
  animation.
\newblock \emph{Proceedings of the ACM on Computer Graphics and Interactive
  Techniques}, 5(1): 1--15.

\bibitem[{Fanelli et~al.(2010{\natexlab{a}})Fanelli, Gall, Romsdorfer, Weise,
  and Gool}]{eth_biwi_00760}
Fanelli, G.; Gall, J.; Romsdorfer, H.; Weise, T.; and Gool, L.~V.
  2010{\natexlab{a}}.
\newblock A 3-D Audio-Visual Corpus of Affective Communication.
\newblock \emph{IEEE Transactions on Multimedia}, 12(6): 591 -- 598.

\bibitem[{Fanelli et~al.(2010{\natexlab{b}})Fanelli, Gall, Romsdorfer, Weise,
  and Van~Gool}]{fanelli20103}
Fanelli, G.; Gall, J.; Romsdorfer, H.; Weise, T.; and Van~Gool, L.
  2010{\natexlab{b}}.
\newblock A 3-d audio-visual corpus of affective communication.
\newblock \emph{IEEE Transactions on Multimedia}, 12(6): 591--598.

\bibitem[{Feng et~al.(2021)Feng, Feng, Black, and Bolkart}]{DECA:Siggraph2021}
Feng, Y.; Feng, H.; Black, M.~J.; and Bolkart, T. 2021.
\newblock Learning an Animatable Detailed {3D} Face Model from In-The-Wild
  Images.
\newblock volume~40.

\bibitem[{Ghosh et~al.(2020)Ghosh, Gupta, Uziel, Ranjan, Black, and
  Bolkart}]{ghosh2020gif}
Ghosh, P.; Gupta, P.~S.; Uziel, R.; Ranjan, A.; Black, M.~J.; and Bolkart, T.
  2020.
\newblock GIF: Generative interpretable faces.
\newblock In \emph{2020 International Conference on 3D Vision (3DV)}, 868--878.
  IEEE.

\bibitem[{Guo et~al.(2020)Guo, Zuo, Wang, Zou, Sun, Deng, Gong, and
  Cheng}]{guo2020action2motion}
Guo, C.; Zuo, X.; Wang, S.; Zou, S.; Sun, Q.; Deng, A.; Gong, M.; and Cheng, L.
  2020.
\newblock Action2motion: Conditioned generation of 3d human motions.
\newblock In \emph{Proceedings of the 28th ACM International Conference on
  Multimedia}, 2021--2029.

\bibitem[{Guo et~al.(2021)Guo, Chen, Liang, Liu, Bao, and Zhang}]{guo2021ad}
Guo, Y.; Chen, K.; Liang, S.; Liu, Y.-J.; Bao, H.; and Zhang, J. 2021.
\newblock Ad-nerf: Audio driven neural radiance fields for talking head
  synthesis.
\newblock In \emph{Proceedings of the IEEE/CVF International Conference on
  Computer Vision}, 5784--5794.

\bibitem[{Habibie et~al.(2021)Habibie, Xu, Mehta, Liu, Seidel, Pons-Moll,
  Elgharib, and Theobalt}]{habibie2021learning}
Habibie, I.; Xu, W.; Mehta, D.; Liu, L.; Seidel, H.-P.; Pons-Moll, G.;
  Elgharib, M.; and Theobalt, C. 2021.
\newblock Learning speech-driven 3d conversational gestures from video.
\newblock In \emph{Proceedings of the 21st ACM International Conference on
  Intelligent Virtual Agents}, 101--108.

\bibitem[{Ho et~al.(2022{\natexlab{a}})Ho, Chan, Saharia, Whang, Gao,
  Gritsenko, Kingma, Poole, Norouzi, Fleet et~al.}]{ho2022imagen}
Ho, J.; Chan, W.; Saharia, C.; Whang, J.; Gao, R.; Gritsenko, A.; Kingma,
  D.~P.; Poole, B.; Norouzi, M.; Fleet, D.~J.; et~al. 2022{\natexlab{a}}.
\newblock Imagen video: High definition video generation with diffusion models.
\newblock \emph{arXiv preprint arXiv:2210.02303}.

\bibitem[{Ho, Jain, and Abbeel(2020)}]{ho2020denoising}
Ho, J.; Jain, A.; and Abbeel, P. 2020.
\newblock Denoising diffusion probabilistic models.
\newblock \emph{Advances in Neural Information Processing Systems}, 33:
  6840--6851.

\bibitem[{Ho and Salimans(2022)}]{ho2022classifier}
Ho, J.; and Salimans, T. 2022.
\newblock Classifier-free diffusion guidance.
\newblock \emph{arXiv preprint arXiv:2207.12598}.

\bibitem[{Ho et~al.(2022{\natexlab{b}})Ho, Salimans, Gritsenko, Chan, Norouzi,
  and Fleet}]{ho2022video}
Ho, J.; Salimans, T.; Gritsenko, A.; Chan, W.; Norouzi, M.; and Fleet, D.~J.
  2022{\natexlab{b}}.
\newblock Video diffusion models.
\newblock \emph{arXiv preprint arXiv:2204.03458}.

\bibitem[{Hong et~al.(2022)Hong, Zhang, Shen, and Xu}]{hong2022depth}
Hong, F.-T.; Zhang, L.; Shen, L.; and Xu, D. 2022.
\newblock Depth-aware generative adversarial network for talking head video
  generation.
\newblock In \emph{Proceedings of the IEEE/CVF Conference on Computer Vision
  and Pattern Recognition}, 3397--3406.

\bibitem[{Hsu et~al.(2021)Hsu, Bolte, Tsai, Lakhotia, Salakhutdinov, and
  Mohamed}]{hsu2021hubert}
Hsu, W.-N.; Bolte, B.; Tsai, Y.-H.~H.; Lakhotia, K.; Salakhutdinov, R.; and
  Mohamed, A. 2021.
\newblock Hubert: Self-supervised speech representation learning by masked
  prediction of hidden units.
\newblock \emph{IEEE/ACM Transactions on Audio, Speech, and Language
  Processing}, 29: 3451--3460.

\bibitem[{Karras et~al.(2017)Karras, Aila, Laine, Herva, and
  Lehtinen}]{karras2017audio}
Karras, T.; Aila, T.; Laine, S.; Herva, A.; and Lehtinen, J. 2017.
\newblock Audio-driven facial animation by joint end-to-end learning of pose
  and emotion.
\newblock \emph{ACM Transactions on Graphics (TOG)}, 36(4): 1--12.

\bibitem[{Liang et~al.(2022)Liang, Pan, Guo, Zhou, Hong, Han, Han, Liu, Ding,
  and Wang}]{liang2022expressive}
Liang, B.; Pan, Y.; Guo, Z.; Zhou, H.; Hong, Z.; Han, X.; Han, J.; Liu, J.;
  Ding, E.; and Wang, J. 2022.
\newblock Expressive talking head generation with granular audio-visual
  control.
\newblock In \emph{Proceedings of the IEEE/CVF Conference on Computer Vision
  and Pattern Recognition}, 3387--3396.

\bibitem[{Loper et~al.(2015)Loper, Mahmood, Romero, Pons-Moll, and
  Black}]{loper2015smpl}
Loper, M.; Mahmood, N.; Romero, J.; Pons-Moll, G.; and Black, M.~J. 2015.
\newblock SMPL: A skinned multi-person linear model.
\newblock \emph{ACM transactions on graphics (TOG)}, 34(6): 1--16.

\bibitem[{Ma, Bai, and Zhou(2022)}]{ma2022pretrained}
Ma, J.; Bai, S.; and Zhou, C. 2022.
\newblock Pretrained Diffusion Models for Unified Human Motion Synthesis.
\newblock \emph{arXiv preprint arXiv:2212.02837}.

\bibitem[{Ma et~al.(2023)Ma, Wang, Hu, Fan, Lv, Ding, Deng, and
  Yu}]{ma2023styletalk}
Ma, Y.; Wang, S.; Hu, Z.; Fan, C.; Lv, T.; Ding, Y.; Deng, Z.; and Yu, X. 2023.
\newblock StyleTalk: One-shot Talking Head Generation with Controllable
  Speaking Styles.
\newblock \emph{arXiv preprint arXiv:2301.01081}.

\bibitem[{Nagrani, Chung, and Zisserman(2017)}]{nagrani2017voxceleb}
Nagrani, A.; Chung, J.~S.; and Zisserman, A. 2017.
\newblock Voxceleb: a large-scale speaker identification dataset.
\newblock \emph{arXiv preprint arXiv:1706.08612}.

\bibitem[{Nichol et~al.(2021)Nichol, Dhariwal, Ramesh, Shyam, Mishkin, McGrew,
  Sutskever, and Chen}]{nichol2021glide}
Nichol, A.; Dhariwal, P.; Ramesh, A.; Shyam, P.; Mishkin, P.; McGrew, B.;
  Sutskever, I.; and Chen, M. 2021.
\newblock Glide: Towards photorealistic image generation and editing with
  text-guided diffusion models.
\newblock \emph{arXiv preprint arXiv:2112.10741}.

\bibitem[{Park et~al.(2022)Park, Kim, Hong, Choi, and
  Ro}]{park2022synctalkface}
Park, S.~J.; Kim, M.; Hong, J.; Choi, J.; and Ro, Y.~M. 2022.
\newblock Synctalkface: Talking face generation with precise lip-syncing via
  audio-lip memory.
\newblock In \emph{Proceedings of the AAAI Conference on Artificial
  Intelligence}, volume~36, 2062--2070.

\bibitem[{Paysan et~al.(2009)Paysan, Knothe, Amberg, Romdhani, and
  Vetter}]{paysan20093d}
Paysan, P.; Knothe, R.; Amberg, B.; Romdhani, S.; and Vetter, T. 2009.
\newblock A 3D face model for pose and illumination invariant face recognition.
\newblock In \emph{2009 sixth IEEE international conference on advanced video
  and signal based surveillance}, 296--301. Ieee.

\bibitem[{Pham, Wang, and Pavlovic(2018)}]{pham2018end}
Pham, H.~X.; Wang, Y.; and Pavlovic, V. 2018.
\newblock End-to-end learning for 3d facial animation from speech.
\newblock In \emph{Proceedings of the 20th ACM International Conference on
  Multimodal Interaction}, 361--365.

\bibitem[{Prajwal et~al.(2020)Prajwal, Mukhopadhyay, Namboodiri, and
  Jawahar}]{prajwal2020lip}
Prajwal, K.; Mukhopadhyay, R.; Namboodiri, V.~P.; and Jawahar, C. 2020.
\newblock A lip sync expert is all you need for speech to lip generation in the
  wild.
\newblock In \emph{Proceedings of the 28th ACM International Conference on
  Multimedia}, 484--492.

\bibitem[{Ramesh et~al.(2022)Ramesh, Dhariwal, Nichol, Chu, and
  Chen}]{ramesh2022hierarchical}
Ramesh, A.; Dhariwal, P.; Nichol, A.; Chu, C.; and Chen, M. 2022.
\newblock Hierarchical text-conditional image generation with clip latents.
\newblock \emph{arXiv preprint arXiv:2204.06125}.

\bibitem[{Ranjan et~al.(2018)Ranjan, Bolkart, Sanyal, and
  Black}]{ranjan2018generating}
Ranjan, A.; Bolkart, T.; Sanyal, S.; and Black, M.~J. 2018.
\newblock Generating 3D faces using convolutional mesh autoencoders.
\newblock In \emph{Proceedings of the European conference on computer vision
  (ECCV)}, 704--720.

\bibitem[{Richard et~al.(2021{\natexlab{a}})Richard, Lea, Ma, Gall, De~la
  Torre, and Sheikh}]{richard2021audio}
Richard, A.; Lea, C.; Ma, S.; Gall, J.; De~la Torre, F.; and Sheikh, Y.
  2021{\natexlab{a}}.
\newblock Audio-and gaze-driven facial animation of codec avatars.
\newblock In \emph{Proceedings of the IEEE/CVF winter conference on
  applications of computer vision}, 41--50.

\bibitem[{Richard et~al.(2021{\natexlab{b}})Richard, Zollh{\"o}fer, Wen, De~la
  Torre, and Sheikh}]{richard2021meshtalk}
Richard, A.; Zollh{\"o}fer, M.; Wen, Y.; De~la Torre, F.; and Sheikh, Y.
  2021{\natexlab{b}}.
\newblock Meshtalk: 3d face animation from speech using cross-modality
  disentanglement.
\newblock In \emph{Proceedings of the IEEE/CVF International Conference on
  Computer Vision}, 1173--1182.

\bibitem[{Rombach et~al.(2022)Rombach, Blattmann, Lorenz, Esser, and
  Ommer}]{rombach2022high}
Rombach, R.; Blattmann, A.; Lorenz, D.; Esser, P.; and Ommer, B. 2022.
\newblock High-resolution image synthesis with latent diffusion models.
\newblock In \emph{Proceedings of the IEEE/CVF Conference on Computer Vision
  and Pattern Recognition}, 10684--10695.

\bibitem[{Saharia et~al.(2022)Saharia, Chan, Saxena, Li, Whang, Denton,
  Ghasemipour, Gontijo~Lopes, Karagol~Ayan, Salimans
  et~al.}]{saharia2022photorealistic}
Saharia, C.; Chan, W.; Saxena, S.; Li, L.; Whang, J.; Denton, E.~L.;
  Ghasemipour, K.; Gontijo~Lopes, R.; Karagol~Ayan, B.; Salimans, T.; et~al.
  2022.
\newblock Photorealistic text-to-image diffusion models with deep language
  understanding.
\newblock \emph{Advances in Neural Information Processing Systems}, 35:
  36479--36494.

\bibitem[{Sanyal et~al.(2019)Sanyal, Bolkart, Feng, and
  Black}]{sanyal2019learning}
Sanyal, S.; Bolkart, T.; Feng, H.; and Black, M.~J. 2019.
\newblock Learning to regress 3D face shape and expression from an image
  without 3D supervision.
\newblock In \emph{Proceedings of the IEEE/CVF Conference on Computer Vision
  and Pattern Recognition}, 7763--7772.

\bibitem[{Savran et~al.(2008)Savran, Aly{\"u}z, Dibeklio{\u{g}}lu,
  {\c{C}}eliktutan, G{\"o}kberk, Sankur, and Akarun}]{savran2008bosphorus}
Savran, A.; Aly{\"u}z, N.; Dibeklio{\u{g}}lu, H.; {\c{C}}eliktutan, O.;
  G{\"o}kberk, B.; Sankur, B.; and Akarun, L. 2008.
\newblock Bosphorus database for 3D face analysis.
\newblock In \emph{Biometrics and Identity Management: First European Workshop,
  BIOID 2008, Roskilde, Denmark, May 7-9, 2008. Revised Selected Papers 1},
  47--56. Springer.

\bibitem[{Shen et~al.(2023)Shen, Zhao, Meng, Li, Zhu, Zhou, and
  Lu}]{shen2023difftalk}
Shen, S.; Zhao, W.; Meng, Z.; Li, W.; Zhu, Z.; Zhou, J.; and Lu, J. 2023.
\newblock DiffTalk: Crafting Diffusion Models for Generalized Talking Head
  Synthesis.
\newblock \emph{arXiv preprint arXiv:2301.03786}.

\bibitem[{Sohl-Dickstein et~al.(2015)Sohl-Dickstein, Weiss, Maheswaranathan,
  and Ganguli}]{sohl2015deep}
Sohl-Dickstein, J.; Weiss, E.; Maheswaranathan, N.; and Ganguli, S. 2015.
\newblock Deep unsupervised learning using nonequilibrium thermodynamics.
\newblock In \emph{International Conference on Machine Learning}, 2256--2265.
  PMLR.

\bibitem[{Song, Meng, and Ermon(2020)}]{song2020denoising}
Song, J.; Meng, C.; and Ermon, S. 2020.
\newblock Denoising diffusion implicit models.
\newblock \emph{arXiv preprint arXiv:2010.02502}.

\bibitem[{Stypu{\l}kowski et~al.(2023)Stypu{\l}kowski, Vougioukas, He,
  Zi{\k{e}}ba, Petridis, and Pantic}]{stypulkowski2023diffused}
Stypu{\l}kowski, M.; Vougioukas, K.; He, S.; Zi{\k{e}}ba, M.; Petridis, S.; and
  Pantic, M. 2023.
\newblock Diffused Heads: Diffusion Models Beat GANs on Talking-Face
  Generation.
\newblock \emph{arXiv preprint arXiv:2301.03396}.

\bibitem[{Tang et~al.(2022)Tang, He, Tan, Ling, Li, Zhao, Song, and
  Bian}]{tang2022memories}
Tang, A.; He, T.; Tan, X.; Ling, J.; Li, R.; Zhao, S.; Song, L.; and Bian, J.
  2022.
\newblock Memories are One-to-Many Mapping Alleviators in Talking Face
  Generation.
\newblock \emph{arXiv preprint arXiv:2212.05005}.

\bibitem[{Taylor et~al.(2017)Taylor, Kim, Yue, Mahler, Krahe, Rodriguez,
  Hodgins, and Matthews}]{taylor2017deep}
Taylor, S.; Kim, T.; Yue, Y.; Mahler, M.; Krahe, J.; Rodriguez, A.~G.; Hodgins,
  J.; and Matthews, I. 2017.
\newblock A deep learning approach for generalized speech animation.
\newblock \emph{ACM Transactions On Graphics (TOG)}, 36(4): 1--11.

\bibitem[{Tevet et~al.(2022)Tevet, Raab, Gordon, Shafir, Cohen-Or, and
  Bermano}]{tevet2022human}
Tevet, G.; Raab, S.; Gordon, B.; Shafir, Y.; Cohen-Or, D.; and Bermano, A.~H.
  2022.
\newblock Human motion diffusion model.
\newblock \emph{arXiv preprint arXiv:2209.14916}.

\bibitem[{Van Den~Oord, Vinyals et~al.(2017)}]{van2017neural}
Van Den~Oord, A.; Vinyals, O.; et~al. 2017.
\newblock Neural discrete representation learning.
\newblock \emph{Advances in neural information processing systems}, 30.

\bibitem[{Vaswani et~al.(2017)Vaswani, Shazeer, Parmar, Uszkoreit, Jones,
  Gomez, Kaiser, and Polosukhin}]{vaswani2017attention}
Vaswani, A.; Shazeer, N.; Parmar, N.; Uszkoreit, J.; Jones, L.; Gomez, A.~N.;
  Kaiser, {\L}.; and Polosukhin, I. 2017.
\newblock Attention is all you need.
\newblock \emph{Advances in neural information processing systems}, 30.

\bibitem[{Wang et~al.(2020)Wang, Wu, Song, Yang, Wu, Qian, He, Qiao, and
  Loy}]{wang2020mead}
Wang, K.; Wu, Q.; Song, L.; Yang, Z.; Wu, W.; Qian, C.; He, R.; Qiao, Y.; and
  Loy, C.~C. 2020.
\newblock Mead: A large-scale audio-visual dataset for emotional talking-face
  generation.
\newblock In \emph{Computer Vision--ECCV 2020: 16th European Conference,
  Glasgow, UK, August 23--28, 2020, Proceedings, Part XXI}, 700--717. Springer.

\bibitem[{Wang, Fan, and Xia(2021)}]{wang20213d}
Wang, Q.; Fan, Z.; and Xia, S. 2021.
\newblock 3d-talkemo: Learning to synthesize 3d emotional talking head.
\newblock \emph{arXiv preprint arXiv:2104.12051}.

\bibitem[{Wang et~al.(2021)Wang, Li, Ding, Fan, and Yu}]{wang2021audio2head}
Wang, S.; Li, L.; Ding, Y.; Fan, C.; and Yu, X. 2021.
\newblock Audio2head: Audio-driven one-shot talking-head generation with
  natural head motion.
\newblock \emph{arXiv preprint arXiv:2107.09293}.

\bibitem[{Wu et~al.(2021)Wu, Jia, Wang, Dou, Duan, and Deng}]{wu2021imitating}
Wu, H.; Jia, J.; Wang, H.; Dou, Y.; Duan, C.; and Deng, Q. 2021.
\newblock Imitating arbitrary talking style for realistic audio-driven talking
  face synthesis.
\newblock In \emph{Proceedings of the 29th ACM International Conference on
  Multimedia}, 1478--1486.

\bibitem[{Wuu et~al.(2022)Wuu, Zheng, Ardisson, Bali, Belko, Brockmeyer, Evans,
  Godisart, Ha, Hypes, Koska, Krenn, Lombardi, Luo, McPhail, Millerschoen,
  Perdoch, Pitts, Richard, Saragih, Saragih, Shiratori, Simon, Stewart,
  Trimble, Weng, Whitewolf, Wu, Yu, and Sheikh}]{wuu2022multiface}
Wuu, C.-h.; Zheng, N.; Ardisson, S.; Bali, R.; Belko, D.; Brockmeyer, E.;
  Evans, L.; Godisart, T.; Ha, H.; Hypes, A.; Koska, T.; Krenn, S.; Lombardi,
  S.; Luo, X.; McPhail, K.; Millerschoen, L.; Perdoch, M.; Pitts, M.; Richard,
  A.; Saragih, J.; Saragih, J.; Shiratori, T.; Simon, T.; Stewart, M.; Trimble,
  A.; Weng, X.; Whitewolf, D.; Wu, C.; Yu, S.-I.; and Sheikh, Y. 2022.
\newblock Multiface: A Dataset for Neural Face Rendering.
\newblock In \emph{arXiv}.

\bibitem[{Xing et~al.(2023)Xing, Xia, Zhang, Cun, Wang, and
  Wong}]{xing2023codetalker}
Xing, J.; Xia, M.; Zhang, Y.; Cun, X.; Wang, J.; and Wong, T.-T. 2023.
\newblock CodeTalker: Speech-Driven 3D Facial Animation with Discrete Motion
  Prior.
\newblock In \emph{Proceedings of the IEEE/CVF Conference on Computer Vision
  and Pattern Recognition (CVPR)}.

\bibitem[{Yin et~al.(2006)Yin, Wei, Sun, Wang, and Rosato}]{yin20063d}
Yin, L.; Wei, X.; Sun, Y.; Wang, J.; and Rosato, M.~J. 2006.
\newblock A 3D facial expression database for facial behavior research.
\newblock In \emph{7th international conference on automatic face and gesture
  recognition (FGR06)}, 211--216. IEEE.

\bibitem[{Zhang and Fisher(2019)}]{zhang20193d}
Zhang, J.; and Fisher, R.~B. 2019.
\newblock 3d visual passcode: Speech-driven 3d facial dynamics for
  behaviometrics.
\newblock \emph{Signal processing}, 160: 164--177.

\bibitem[{Zhang et~al.(2022)Zhang, Cai, Pan, Hong, Guo, Yang, and
  Liu}]{zhang2022motiondiffuse}
Zhang, M.; Cai, Z.; Pan, L.; Hong, F.; Guo, X.; Yang, L.; and Liu, Z. 2022.
\newblock Motiondiffuse: Text-driven human motion generation with diffusion
  model.
\newblock \emph{arXiv preprint arXiv:2208.15001}.

\bibitem[{Zhang et~al.(2023)Zhang, Guo, Pan, Cai, Hong, Li, Yang, and
  Liu}]{zhang2023remodiffuse}
Zhang, M.; Guo, X.; Pan, L.; Cai, Z.; Hong, F.; Li, H.; Yang, L.; and Liu, Z.
  2023.
\newblock ReMoDiffuse: Retrieval-Augmented Motion Diffusion Model.
\newblock \emph{arXiv preprint arXiv:2304.01116}.

\bibitem[{Zhang et~al.(2013)Zhang, Yin, Cohn, Canavan, Reale, Horowitz, and
  Liu}]{zhang2013high}
Zhang, X.; Yin, L.; Cohn, J.~F.; Canavan, S.; Reale, M.; Horowitz, A.; and Liu,
  P. 2013.
\newblock A high-resolution spontaneous 3d dynamic facial expression database.
\newblock In \emph{2013 10th IEEE international conference and workshops on
  automatic face and gesture recognition (FG)}, 1--6. IEEE.

\bibitem[{Zhang et~al.(2014)Zhang, Yin, Cohn, Canavan, Reale, Horowitz, Liu,
  and Girard}]{zhang2014bp4d}
Zhang, X.; Yin, L.; Cohn, J.~F.; Canavan, S.; Reale, M.; Horowitz, A.; Liu, P.;
  and Girard, J.~M. 2014.
\newblock Bp4d-spontaneous: a high-resolution spontaneous 3d dynamic facial
  expression database.
\newblock \emph{Image and Vision Computing}, 32(10): 692--706.

\bibitem[{Zhang et~al.(2016)Zhang, Girard, Wu, Zhang, Liu, Ciftci, Canavan,
  Reale, Horowitz, Yang et~al.}]{zhang2016multimodal}
Zhang, Z.; Girard, J.~M.; Wu, Y.; Zhang, X.; Liu, P.; Ciftci, U.; Canavan, S.;
  Reale, M.; Horowitz, A.; Yang, H.; et~al. 2016.
\newblock Multimodal spontaneous emotion corpus for human behavior analysis.
\newblock In \emph{Proceedings of the IEEE conference on computer vision and
  pattern recognition}, 3438--3446.

\bibitem[{Zhang et~al.(2021)Zhang, Li, Ding, and Fan}]{zhang2021flow}
Zhang, Z.; Li, L.; Ding, Y.; and Fan, C. 2021.
\newblock Flow-guided one-shot talking face generation with a high-resolution
  audio-visual dataset.
\newblock In \emph{Proceedings of the IEEE/CVF Conference on Computer Vision
  and Pattern Recognition}, 3661--3670.

\bibitem[{Zhou et~al.(2019)Zhou, Liu, Liu, Luo, and Wang}]{zhou2019talking}
Zhou, H.; Liu, Y.; Liu, Z.; Luo, P.; and Wang, X. 2019.
\newblock Talking face generation by adversarially disentangled audio-visual
  representation.
\newblock In \emph{Proceedings of the AAAI conference on artificial
  intelligence}, volume~33, 9299--9306.

\bibitem[{Zhou et~al.(2021)Zhou, Sun, Wu, Loy, Wang, and Liu}]{zhou2021pose}
Zhou, H.; Sun, Y.; Wu, W.; Loy, C.~C.; Wang, X.; and Liu, Z. 2021.
\newblock Pose-controllable talking face generation by implicitly modularized
  audio-visual representation.
\newblock In \emph{Proceedings of the IEEE/CVF conference on computer vision
  and pattern recognition}, 4176--4186.

\bibitem[{Zhou et~al.(2018)Zhou, Xu, Landreth, Kalogerakis, Maji, and
  Singh}]{zhou2018visemenet}
Zhou, Y.; Xu, Z.; Landreth, C.; Kalogerakis, E.; Maji, S.; and Singh, K. 2018.
\newblock Visemenet: Audio-driven animator-centric speech animation.
\newblock \emph{ACM Transactions on Graphics (TOG)}, 37(4): 1--10.

\bibitem[{Zhu et~al.(2021)Zhu, Yang, Guo, Zhang, Wang, Huang, Shen, Yang, and
  Cao}]{zhu2021facescape}
Zhu, H.; Yang, H.; Guo, L.; Zhang, Y.; Wang, Y.; Huang, M.; Shen, Q.; Yang, R.;
  and Cao, X. 2021.
\newblock FacesCape: 3D facial dataset and benchmark for single-view 3D face
  reconstruction.
\newblock \emph{arXiv preprint arXiv:2111.01082}.

\end{thebibliography}

\end{document}